\documentclass[journal]{IEEEtran}
\ifCLASSINFOpdf
\else
\fi
\usepackage{graphicx}
\usepackage{booktabs}
\usepackage{microtype}
\usepackage{float}
\usepackage{bbding}
\usepackage{amsmath}
\usepackage{amsfonts}
\usepackage{amssymb}
\usepackage{dblfloatfix}
\newcommand{\etal}{\textit{et al.}}

\usepackage[pagebackref=false,breaklinks=true,letterpaper=true,colorlinks,linkcolor=blue,citecolor=blue,urlcolor=blue]{hyperref}

\begin{document}
\title{Bio-SFT: Asymmetric Cortical Guidance and Retinal Adaptation for Robust HDR Reconstruction}

\author{Tingyu Cheng, Ting Zhang,
        Chongyi Li,~\IEEEmembership{Senior Member,~IEEE,}
        Zhaoqing Pan,~\IEEEmembership{Senior Member,~IEEE,}\\
        and Tiesong Zhao,~\IEEEmembership{Senior Member,~IEEE}% <-this % stops a space

\thanks{This work is supportedby the National Science Foundationof China (Grant No.62571131). \textit{(Corresponding author: Tiesong Zhao.)}}
\thanks{Tingyu Cheng, Ting Zhang and Tiesong Zhao are with the Fujian Key Laboratory of Intelligent Processing and Wireless Transmission of Media Information, College of Physics and Information Engineering, Fuzhou University, Fujian 350108, China (e-mail: \{241127011, 231110033, t.zhao\}@fzu.edu.cn).}% <-this % stops a space
\thanks{Chongyi Li is with the School of Computer Science, Nankai University,Tianjin 300071, China (e-mail: lichongyi@nankai.edu.cn).}% <-this % stops a space
\thanks{Zhaoqing Pan is with theSchool of Electrical and Information Engineering, Tianjin University, Tianjin300072, China (e-mail: zqpan3-c@my.cityu.edu.hk).} 
}


\markboth{Journal of \LaTeX\ Class Files,~Vol.~X, No.~X, June~2026}%
{Shell \MakeLowercase{\textit{et al.}}: Bio-SFT: Asymmetric Cortical Guidance and Retinal Adaptation for Robust HDR Reconstruction}



\maketitle


\begin{abstract}
Recovering high dynamic range (HDR) radiance from a single standard dynamic range (SDR) image is highly ill-posed. Extreme luminance variation and severe quantization in dark regions make accurate reconstruction challenging, often leading to visual artifacts and color distortions. To address this problem, we propose Bio-SFT, a bio-inspired spiking frequency transformer for single-image HDR reconstruction. Bio-SFT incorporates three biologically motivated components. First, a learnable Naka--Rushton retinal adaptation frontend stabilizes the input under complex lighting conditions. Second, an explicit Parvo--Magno split introduces asymmetric Parvo-to-Magno guidance, allowing high-frequency structural cues to modulate low-frequency reconstruction. Third, an event-driven SNN hard gating module applies all-or-none spiking to suppress dark-region noise while preserving structural details. The module is trained with a sparsity prior to encourage efficient feature utilization. Built for end-to-end training within a transformer backbone, these lightweight components provide strong parameter efficiency. Experiments on HDRTV1K show that Bio-SFT achieves competitive perceptual quality and consistently improves HDR-VDP-3 and $\Delta E_{ITP}$ while reducing artifact propagation in symmetric guidance pipelines.
\end{abstract}


% \fi
\IEEEpeerreviewmaketitle

\vspace{-2mm}
\section{Introduction}
\label{sec:intro}

\IEEEPARstart{H}{igh} Dynamic Range (HDR) imaging is essential for immersive visual experiences because it preserves scene radiance over the wide luminance range encountered in real-world scenes. The human visual system (HVS) can perceive scenes spanning many orders of magnitude in luminance through adaptive mechanisms in the retina and early visual pathway, whereas consumer cameras and displays are typically limited to Standard Dynamic Range (SDR), which captures only a limited portion of scene radiance. Consequently, single-image HDR (SI-HDR) reconstruction---recovering lost highlight and shadow information from an SDR input---remains a highly ill-posed problem that must balance dynamic range expansion, fidelity, and noise suppression~\cite{le2023single, wu2022litmnet}.

Modern SI-HDR methods predominantly rely on CNNs or transformers, sometimes incorporating explicit imaging priors to improve fidelity~\cite{liu2020single}. Recently, generative models such as diffusion and VQGANs have also been applied to LDR-to-HDR conversion. These methods synthesize photorealistic details and constrain the solution space using real HDR priors~\cite{wang2025lediff, he2024beyond}. In parallel, emerging backward-compatible formats have focused on Gain Map estimation. Such methods employ lightweight MLPs, dual-branch networks, or decomposed diffusion to learn the SDR-to-HDR mapping rather than directly predicting HDR pixels~\cite{canham2025gain, liao2025learning, guan2025hdr}.

\begin{figure}[t]
  \centering
  % Placeholder for the explanatory figure
  \includegraphics[width=\linewidth]{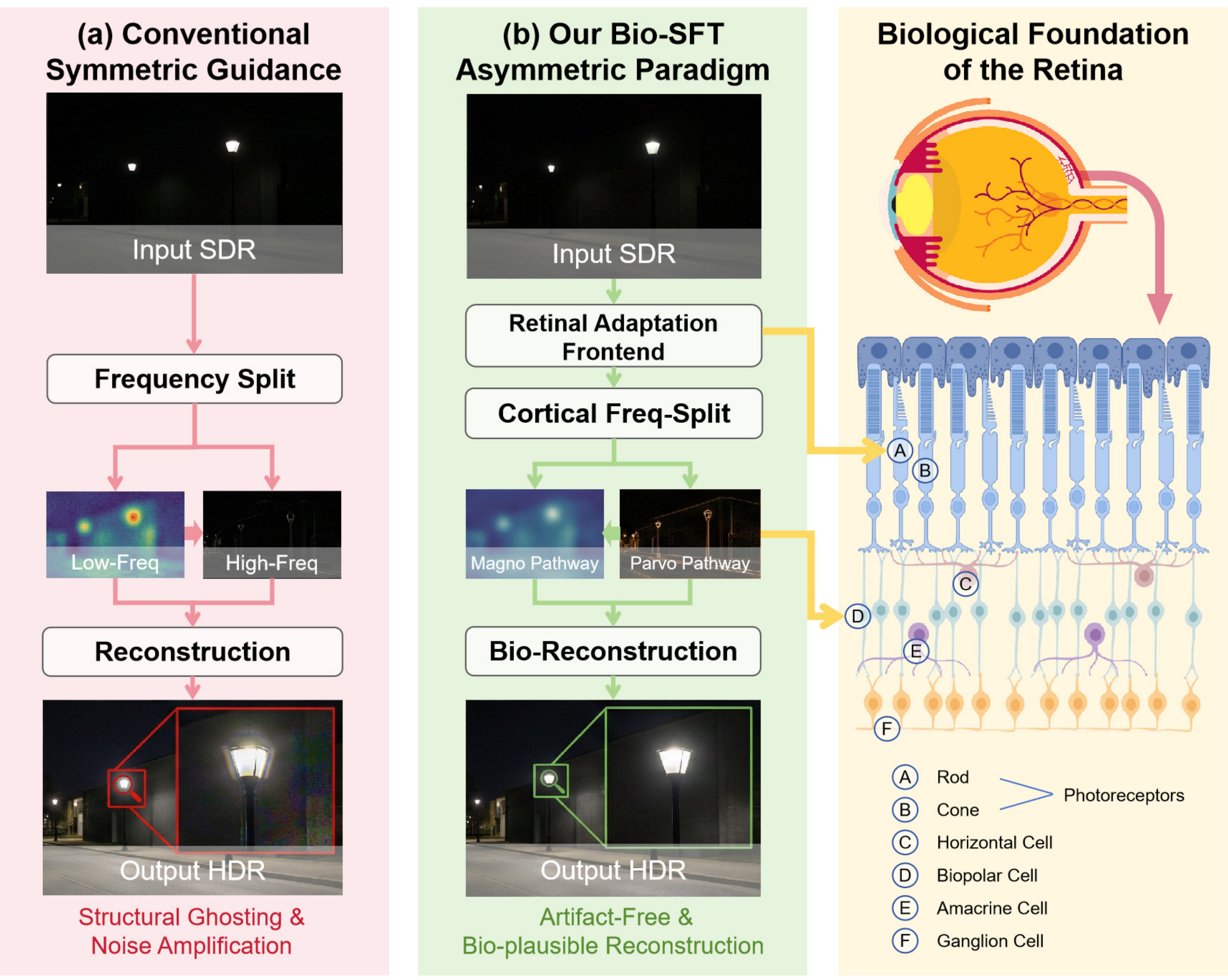}
  \vspace{-6mm}
  \caption{Overview of our bio-inspired HDR framework. (a) Conventional symmetric guidance propagates low-frequency quantization noise into structural details, causing severe ghosting. (b) Bio-SFT enforces asymmetric Parvo-to-Magno guidance for artifact-free reconstruction, structurally grounded in adaptive Photoreceptors (A/B) and edge-sensitive Bipolar Cells (D).}
  \label{fig:teaser}
  \vspace{-7mm}
\end{figure}

However, these approaches typically treat radiance recovery as either a direct mapping problem or a generative task. They rely on static nonlinear operators or symmetric multi-scale fusion rules, which do not explicitly model the adaptive, frequency-selective processing observed in the HVS. As conceptualized in Fig.~\ref{fig:teaser}, prevalent symmetric or ``coarse-to-fine'' guidance strategies exhibit critical vulnerabilities under extreme HDR regimes. While effective in general image restoration tasks, they indiscriminately propagate severe quantization noise residing in underexposed low-frequency backgrounds into high-frequency structural textures, resulting in visible ghosting artifacts and chromatic shifts. To overcome this vulnerability, our Bio-SFT framework draws direct structural inspiration from the physiological mechanisms of the human retina. Specifically, our Retinal Adaptation Frontend emulates the early dynamic-range compression of Photoreceptors (A/B), while the detail-preserving Parvo Pathway reflects the center-surround contrast sensitivity of Bipolar Cells (D). By enforcing a unidirectional, asymmetric Parvo-to-Magno ($H \rightarrow L$) guidance flow, Bio-SFT leverages pristine, high-frequency edge events to calibrate ambiguous low-frequency radiance, achieving artifact-free reconstruction while maintaining strict photometric consistency.

\begin{figure}[htbp]
  \centering
  \includegraphics[width=\linewidth]{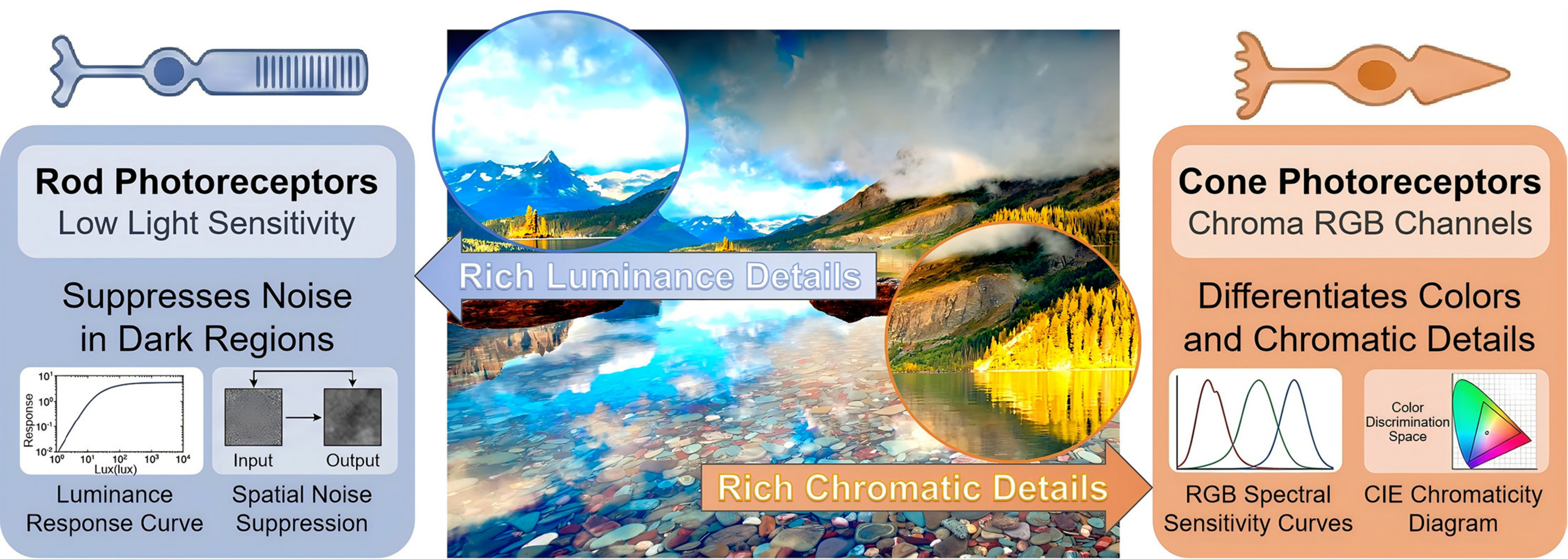}
  \vspace{-5mm}
  \caption{Biological motivation for decoupling signals in HDR imaging. The Human Visual System (HVS) handles extreme dynamic range by splitting stimuli into rod and cone pathways. Inspired by this, our method decouples luminance and chrominance before feature extraction to preserve color fidelity under extreme lighting.}
  \label{fig:motivation}
  \vspace{-2mm}
\end{figure}

Motivated by biological vision, prior work has explored retina-inspired tone mapping and dual-pathway enhancement for dynamic range compression~\cite{zhang2020retina, yang2019biological, xiang2020biological}. As shown in Fig.~\ref{fig:motivation}, the HVS separates visual signals into a luminance pathway, which supports low-light sensitivity, and a chrominance pathway, which supports color fidelity. In addition, cortical processing tends to separate coarse structural information from fine textural details, which helps preserve salient edges. Based on these observations, we design an HDR pipeline with three principles: (a) decouple luminance and chrominance for adaptive preprocessing, (b) split low- and high-frequency features to mimic cortical processing, and (c) apply event-driven hard gating to suppress noise in high-frequency responses.

\vspace{-2mm}

To realize this idea, we propose \textbf{Bio-SFT} (Bio-inspired Spiking Frequency Transformer), a lightweight SI-HDR framework with 0.14M parameters. Bio-SFT integrates physiological visual mechanisms into an end-to-end reconstruction pipeline. It first stabilizes scene radiance through a data-driven \textit{Physiological Retinal Adaptation Frontend}. It then decouples deep features through a \textit{Cortical Pathway Split}. Finally, an \textit{Event-Driven SNN Hard Gating} module and \textit{Asymmetric Parvo-to-Magno Cortical Guidance} ($H \rightarrow L$) work together to suppress noise and guide reconstruction from reliable high-frequency cues to ambiguous low-frequency regions. This design helps reduce the propagation of quantization artifacts and improves reconstruction robustness in dark regions.

Our main contributions are summarized as follows:
\begin{itemize}
  \item \textbf{Learnable Retinal Adaptation:} We develop a Physiological Retinal Adaptation Frontend that reformulates the static Naka--Rushton equation as a spatially adaptive preprocessing layer. It provides robust dynamic-range compression and reduces overexposure before deep feature extraction.
  \item \textbf{SNN-Guided Asymmetric Guidance:} We propose an Asymmetric Cortical Guidance mechanism coupled with Event-Driven SNN Hard Gating. This $H \rightarrow L$ strategy uses the thresholding behavior of spiking neurons to suppress low-amplitude noise and reduce structural crosstalk.
  \item \textbf{Lightweight HDR Reconstruction Framework:} We design Bio-SFT as a unified transformer-based architecture with a sparsity prior. Experiments on HDRTV1K show that Bio-SFT achieves strong perceptual quality, including HDR-VDP-3 and $\Delta E_{ITP}$, while maintaining only 0.14M parameters.
\end{itemize}

The remainder of this paper is organized as follows. Section~\ref{sec:related_work} reviews the related literature on deep learning-based HDR reconstruction, bio-inspired visual computing, and Spiking Neural Networks (SNNs). Section~\ref{sec:method} elaborates on the proposed Bio-SFT framework, detailing the physiological retinal adaptation front-end, the hierarchical cortical backbone, the event-driven SNN hard gating, and the asymmetric Parvo-to-Magno guidance mechanisms. Section~\ref{sec:experiment} presents comprehensive experimental results, including cross-dataset evaluations on the HDRTV1K and SI-HDR benchmarks, 4K computational efficiency analysis, and detailed ablation studies. Section~\ref{sec:discussion} provides a broader discussion of the proposed method. Finally, Section~\ref{sec:conclusion} concludes the paper.

\section{Related Work}
\label{sec:related_work}

\subsection{Deep Learning-based HDR Reconstruction}
\label{subsec:rel_hdr}

It is crucial to distinguish modern broadcast SDR-to-HDR translation from traditional Inverse Tone Mapping (ITM). Traditional ITM treats Low Dynamic Range (LDR) images as display-referred outputs and primarily focuses on reversing Camera Response Function (CRF) degradation. Early deep learning approaches, such as HDR-CNN~\cite{eilertsen2017hdr}, MaskHDR~\cite{santos2020single}, and GAN-based DrTMO~\cite{endo2017deep}, improved highlight hallucination but often lacked explicit radiometric constraints, leading to artifacts or training instability. Subsequent pipeline-based methods like SingleHDR~\cite{liu2020single} and HDRUNet~\cite{chen2021hdrunet} explicitly modeled this camera formation and dequantization process. While improving interpretability, they remain highly sensitive to assumed degradation models and struggle with diverse noise patterns.

In contrast, modern television production strictly adheres to the Rec.2100 standard, encoding HDR in the pixel domain with higher color depths (10/12-bit) and advanced optical-electronic transfer functions (OETF) like the Perceptual Quantizer (PQ). Addressing this specific broadcast standard, Chen \etal~\cite{chen2021new} proposed a three-step SDRTV-to-HDRTV pipeline (global color mapping, local enhancement, and highlight generation), which was later refined to reduce gamut-transition artifacts~\cite{chen2025towards}. Despite their effectiveness, these specialized pipelines are computationally demanding.

Recently, transformers have become dominant in low-level vision, with SwinIR~\cite{liang2021swinir} and Uformer~\cite{wang2022uformer} demonstrating exceptional long-range modeling. Closely related to our work is the Dual Frequency Transformer (DFT)~\cite{xu2024dual}, which separates features into high- and low-frequency components. However, DFT relies on standard Softmax attention. In dark, noise-contaminated SDR regions, such symmetric attention assigns non-negligible weights to quantization artifacts, blurring the separation between useful structures and corrupt signals. This limitation underscores the need for the asymmetric, strict noise-filtering mechanisms proposed in our Bio-SFT.

\begin{figure*}[t]
  \centering
  \includegraphics[width=\textwidth]{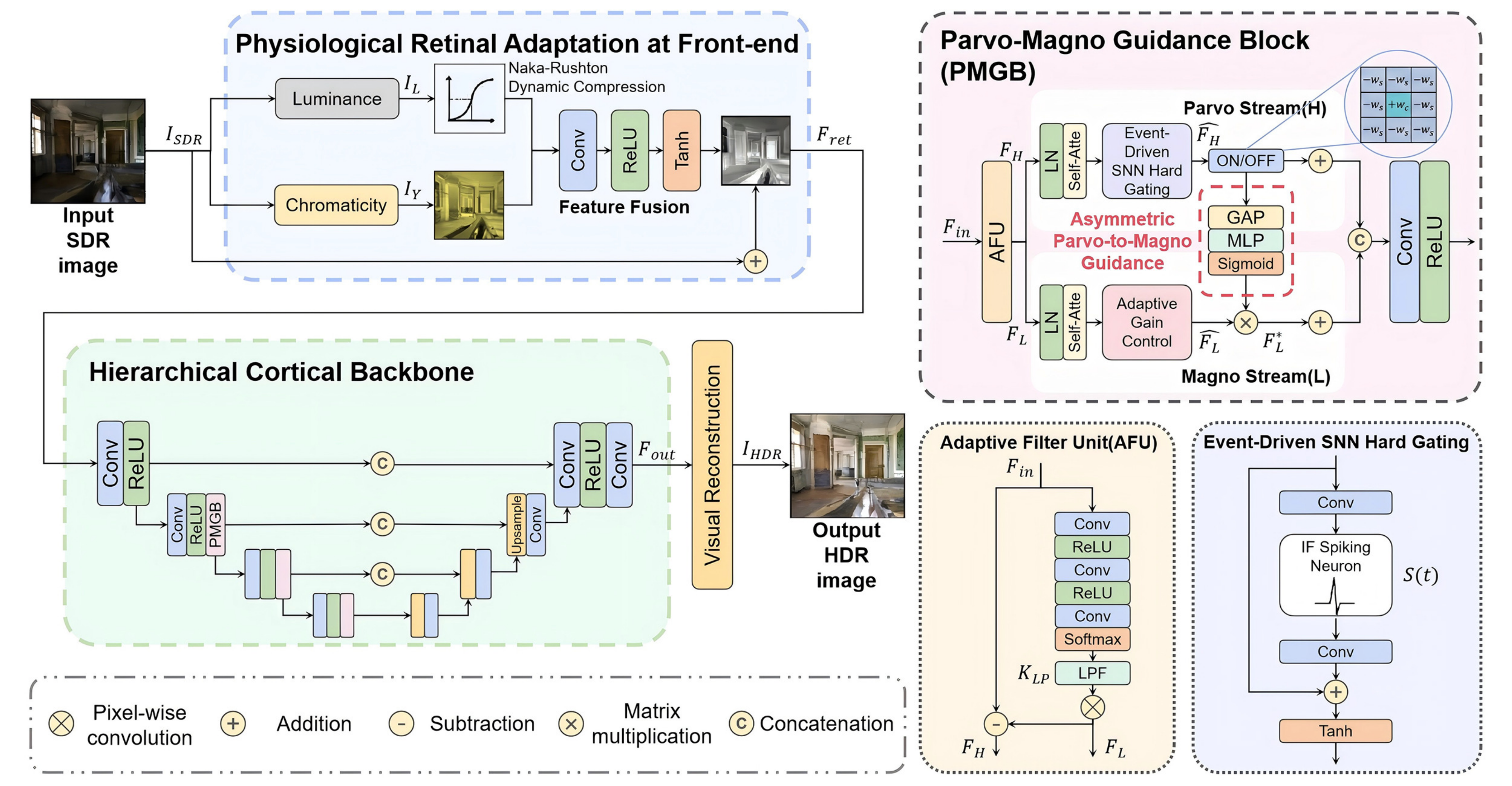}
  \vspace{-5mm}
  \caption{The overall architecture of the Bio-SFT framework. The pipeline aligns with the biological visual system through three sequential stages: (1) a Physiological Retinal Adaptation Front-end, (2) a U-shaped backbone built upon Parvo-Magno Guidance Blocks (PMGBs), and (3) a Visual Reconstruction Module. As detailed in the bottom-right, the PMGB integrates Event-Driven SNN Hard Gating and Asymmetric Parvo-to-Magno Guidance to robustly filter and fuse frequency features.}
  \label{fig:architecture}
  \vspace{-5mm}
\end{figure*}

\subsection{Bio-inspired Visual Computing}
\label{subsec:rel_bio}
Biological vision systems exhibit strong adaptability to a wide range of luminance conditions, which has inspired many computational models for image processing. Classical Retinex theory~\cite{land1971lightness} and its variants~\cite{jobson1997multiscale} decompose images into illumination and reflectance components, and have become important foundations for image enhancement. To model the nonlinear response of photoreceptors, the Naka--Rushton equation~\cite{naka1966s} was introduced. Pu \etal~\cite{pu2017retinal} proposed a retinal adaptation model based on cone and rod channels for HDR compression. Similarly, Zhang \etal~\cite{zhang2016retina} designed a retina-inspired model that simulates adaptive gap junctions and bipolar cells for HDR rendering on LDR displays.

Beyond single-neuron modeling, pathway-level mechanisms have also been explored. Yang \etal~\cite{yang2019biological} developed a two-pathway framework that mimics the Magno- and Parvo-cellular pathways for low-light enhancement. Other works have applied bio-priors to saliency detection~\cite{shen2025retina} and color constancy~\cite{gao2015color}. These methods show the value of biological priors, but many of them rely on hand-crafted parameters, such as a fixed $\sigma$ in the Naka--Rushton model, which limits adaptability to diverse image statistics. In addition, most of these methods were developed for tone mapping or general enhancement rather than the inverse tone mapping problem. In contrast, our work introduces a learnable retinal adaptation frontend that predicts physiological parameters directly from input data.

\vspace{-2mm}
\subsection{Spiking Neural Networks in Computer Vision}
\label{subsec:rel_snn}
Spiking Neural Networks (SNNs) emulate the discrete information processing of biological neurons. Early SNNs relied on unsupervised learning rules such as STDP~\cite{bi1998synaptic}, which limited network depth and scalability. The introduction of surrogate gradients~\cite{neftci2019surrogate} enabled effective training of deep SNNs. More recently, SNNs have been applied to high-level tasks such as classification, with architectures including Spikformer~\cite{zhou2022spikformer} and Spike-driven Transformer~\cite{yao2023spike} combining spikes with self-attention for improved efficiency.

SNNs are also gaining attention in low-level vision~\cite{hopkins2018spiking}. For example, Zhu \etal~\cite{zhu2025high} introduced multi-level spike streams and a Mamba network for dynamic HDR reconstruction. Zhang \etal~\cite{zhang2025Dehaze} proposed a retina-inspired SNN with ON/OFF pathways for efficient dehazing. Although these methods share a bio-inspired motivation, they mainly emphasize computational efficiency or specific restoration tasks. They do not directly address the challenge of HDR reconstruction under severe quantization noise and extreme luminance variation. Moreover, many hybrid networks use SNNs mainly as an alternative to standard arithmetic operations. In contrast, we exploit the all-or-none firing behavior of IF neurons as a hard gating mechanism for suppressing low-amplitude noise.

\vspace{-2mm}
\section{Methodology}
\label{sec:method}

\subsection{Overview of Bio-SFT}
\label{subsec:overview}

As illustrated in Fig.~\ref{fig:architecture}, the network follows a three-stage pipeline closely aligned with the biological visual system: the \textit{Physiological Retinal Adaptation Front-end} (Sec.~\ref{subsec:retina}), the \textit{Parvo-Magno Guidance Block (PMGB)} (Sec.~\ref{subsec:pmgb} and Sec.~\ref{subsec:snn}) embedded within a U-shaped backbone, and the \textit{Visual Reconstruction Module}.

Unlike standard HDR methods that feed raw SDR inputs directly into a deep backbone, Bio-SFT first applies the Physiological Retinal Adaptation Front-end ($\Phi_{retina}$) for early dynamic range compression, simulating adaptive photoreceptors to produce a bio-plausible representation $F_{ret}$.

A shallow convolutional layer then projects $F_{ret}$ into a higher-dimensional feature space. These features are processed by a hierarchical U-shaped architecture, where standard convolutional layers are replaced by our proposed PMGBs. Within each PMGB, features are dynamically decoupled into Parvo (high-frequency) and Magno (low-frequency) streams to enforce an asymmetric modulation. Crucially, an Event-Driven SNN Hard Gating mechanism explicitly traps dark-zone quantization noise before frequency interaction. Finally, the Visual Reconstruction Module maps the deep features back to the spatial image domain, producing the restored HDR image.

\subsection{Physiological Retinal Adaptation at Front-End}
\label{subsec:retina}

\begin{figure}[b]
  \vspace{-3mm}
  \centering
  \includegraphics[width=\linewidth]{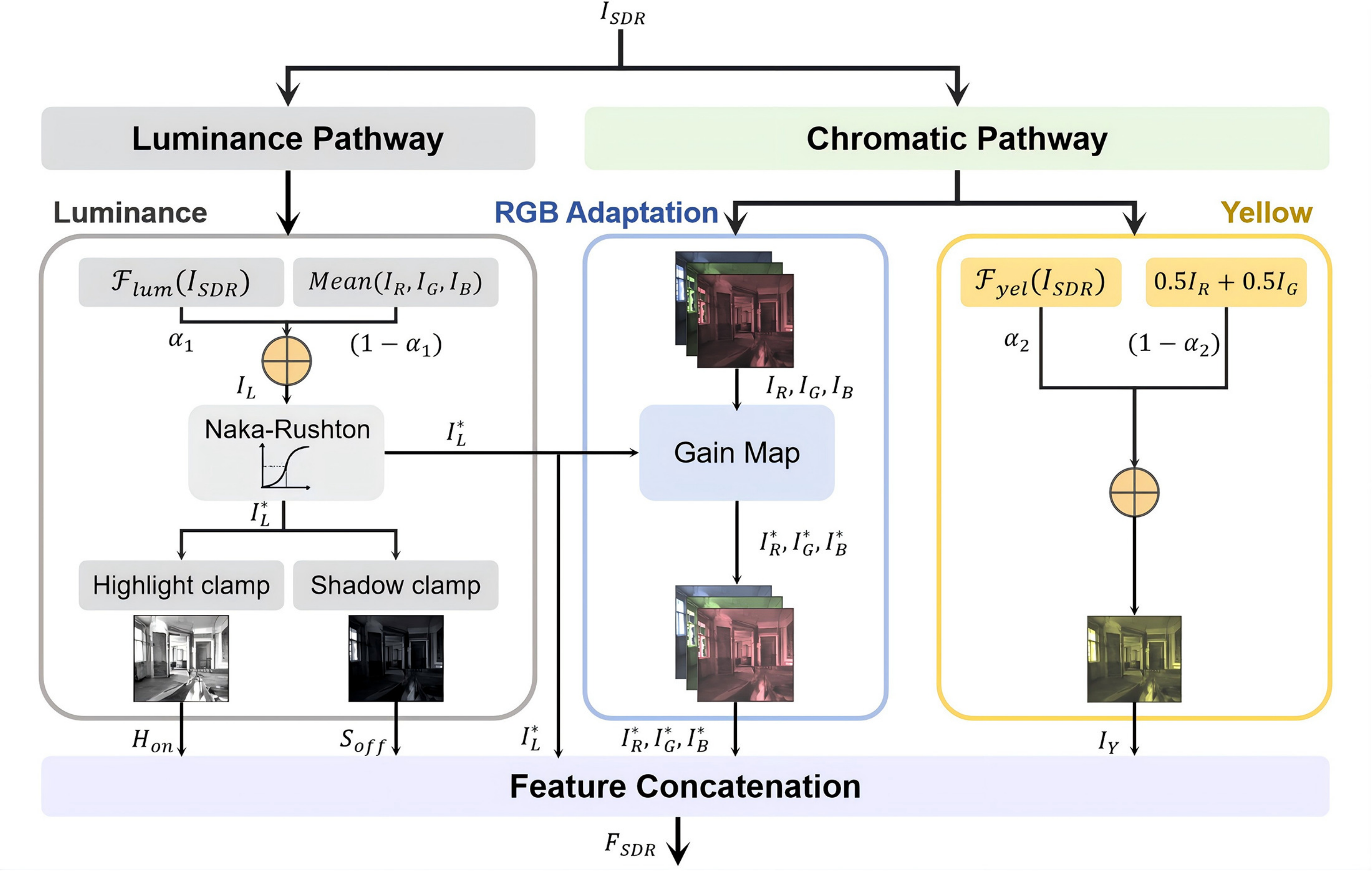}
  \vspace{-6mm}
  \caption{Illustration of the Physiological Retinal Adaptation Front-end. The module produces a 7-channel bio-signal tensor ($I_R^*, I_G^*, I_B^*, I_L^*, I_Y, H_{on}, S_{off}$) by combining learnable Naka–Rushton adaptation with rod/cone pathways.}
  \label{fig:retina_module}
  \vspace{-3mm}
\end{figure}

In the biological visual system, the retina performs early dynamic-range adaptation, enabling the perception of extreme contrast without saturation. Global tone-mapping operators in prior HDR frameworks cannot capture this spatially varying photoreceptor sensitivity. We therefore design a Physiological Retinal Adaptation Front-end (Fig.~\ref{fig:retina_module}) that emulates physiological retinal gain control. Rather than using fixed compression, this module decouples the SDR input into independent luminance and chrominance pathways. By doing so, it learns a nonlinear response curve that increases sensitivity in underexposed regions while compressing saturated highlights prior to deep feature extraction.

To mimic the physiological separation of rod and cone cells, we first decompose the RGB input into a luminance pathway ($I_L$) and a yellow-chromatic pathway ($I_Y$):
\begin{equation}
  \begin{aligned}
  I_L &= \alpha_{1}\,\mathcal{F}_{lum}(I_{SDR}) + (1-\alpha_{1})\,\mathrm{Mean}(I_R,I_G,I_B), \\
  I_Y &= \alpha_{2}\,\mathcal{F}_{yel}(I_{SDR}) + (1-\alpha_{2})\,(0.5 I_R + 0.5 I_G),
  \end{aligned}
  \label{eq:ly_decomp}
\end{equation}
where $\mathcal{F}_{lum}$ and $\mathcal{F}_{yel}$ denote lightweight convolutional feature extractors designed to capture deep luminance and yellow-chromatic representations from the SDR input, respectively, and $\alpha_{1}, \alpha_{2}$ are learnable fusion weights.

\textbf{Luminance Pathway.} 
To process the rod-dominated luminance signal, we reformulate the Naka--Rushton equation as a spatially learnable dynamic compression function, denoted as $R_{nr}(\cdot)$. Applying this function to the luminance channel $I_L$ yields:
\begin{equation}
  R_{nr}= \frac{I_L^{\,n}}{I_L^{\,n} + \sigma^{\,n}},
  \label{eq:naka_rushton}
\end{equation}
with positivity enforced by
\begin{equation}
  n = \mathrm{softplus}(n'), \qquad \sigma = \mathrm{softplus}(\sigma') + \epsilon,
  \label{eq:n_sigma_param}
\end{equation}
where \(n',\sigma'\) are predicted parameter maps and \(\epsilon=10^{-6}\) prevents division by zero. The input $I_L$ is clamped to \([\epsilon, I_{\max}]\) (e.g., \(I_{\max}=10^4\)) before exponentiation.

To improve stability near zero and smoothly interpolate between the linear luminance and its compressed Naka-Rushton representation, we employ a gated residual adaptation:
\begin{equation}
  I_L^* = I_L + g(I_L)\,\big(R_{nr} - I_L\big),
  \label{eq:adaptation_gated}
\end{equation}
where the gating function is defined as $g(I_L) = \operatorname{sigmoid}\big(k\,(I_L - \tau)\big)$, with $k$ and $\tau$ being learnable parameters (default $k=10,\ \tau=0.1$).

Furthermore, we extract ON and OFF event maps from the adapted luminance \(I_L^*\) using differentiable clamps. These are implemented with smooth approximations to preserve stable gradients during end-to-end training:
\begin{equation}
  H_{on} = \mathrm{Clamp}\big(\gamma (I_L^* - \tau_{high}),\,0,\,1\big),
  \label{eq:on_clamp}
\end{equation}
\begin{equation}
  S_{off} = \mathrm{Clamp}\big(\gamma (\tau_{low} - I_L^*),\,0,\,1\big),
  \label{eq:off_clamp}
\end{equation}
where \(\tau_{high},\tau_{low}\) and \(\gamma\) are learnable parameters.

\begin{figure*}[b] 
  \centering
  \includegraphics[width=\textwidth]{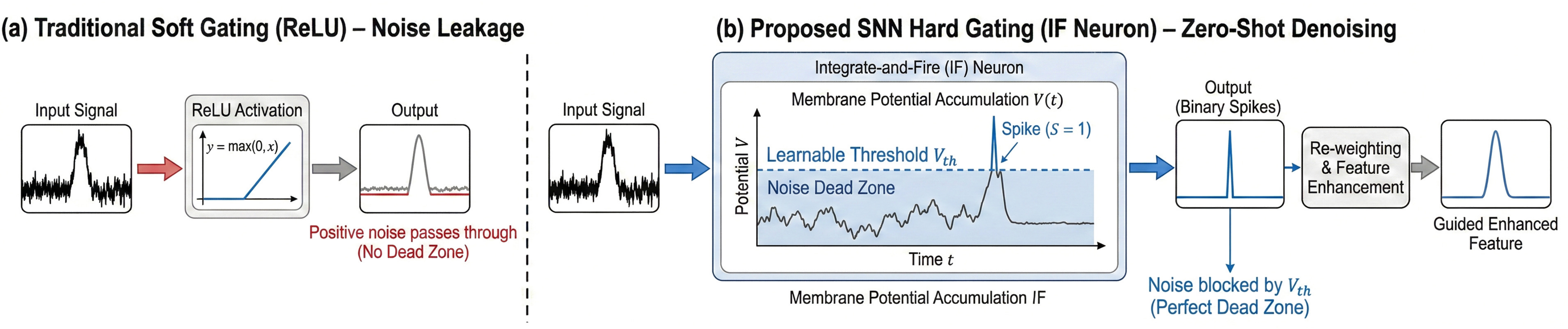} 
  \vspace{-5mm}
  \caption{Mechanistic comparison of gating strategies. (a) Traditional continuous soft gating (e.g., ReLU) indiscriminately leaks low-amplitude quantization noise due to the absence of a dead zone. (b) Our proposed Event-Driven SNN Hard Gating leverages the Integrate-and-Fire (IF) neuron's membrane potential accumulation $V(t)$. Spatially inconsistent noise fails to reach the learnable threshold $V_{th}$ and is permanently blocked within the ``Noise Dead Zone'', enabling robust zero-shot denoising while preserving sharp structural spikes.}
  \label{fig:snn_module}
  \vspace{-3mm}
\end{figure*}

\textbf{Chrominance Pathway.} 
To preserve color fidelity and prevent chromatic distortion during severe dynamic range compression, the initial chroma channels $I_c \in \{I_R, I_G, I_B\}$ are adaptively modulated by the local luminance adaptation ratio. Inspired by the color constancy mechanism in the human visual system, the adapted chroma channels $I_c^*$ are formulated via a spatially-adaptive gating process:
\begin{equation}
  I_c^* = \operatorname{clip}\Big(I_c\big(1 + w(x)\cdot (\tfrac{R_{nr}}{I_L + \epsilon} - 1)\big),\,0,\,I_{\text{clip}}\Big).
  \label{eq:adapt_rgb}
\end{equation}
Here, $\tfrac{R_{nr}}{I_L + \epsilon}$ represents the localized luminance modulation factor governed by the Naka-Rushton operational response. The residual term $(\tfrac{R_{nr}}{I_L + \epsilon} - 1)$ captures the relative scaling deviation of the luminance pathway. This is further regulated by a learnable, per-pixel modulation amplitude $w(x)$ (spatially initialized near zero) to flexibly tune the chrominance gain. Finally, a $\operatorname{clip}(\cdot)$ operation bounds the output range within $[0, I_{\text{clip}}]$, thereby suppressing out-of-bound chromatic artifacts and stabilizing gradient propagation.

Finally, the front-end output is constructed by concatenating these decoupled and enhanced features into a comprehensive retinal tensor:
\begin{equation}
  F_{ret} = \mathrm{Concat}\big([I_R^*, I_G^*, I_B^*, I_L^*, I_Y, H_{on}, S_{off}]\big).
  \label{eq:fret}
\end{equation}

\subsection{Parvo-Magno Guidance Block}
\label{subsec:pmgb}

After retinal preprocessing, visual information propagates to the primary visual cortex (V1). To operationalize this, we process the adapted deep features through a hierarchical U-shaped backbone. The core computational unit deployed at each scale of this architecture is the Parvo-Magno Guidance Block (PMGB), designed to mimic the biological separation of the Parvo-cellular (P-pathway) for high-frequency details and the Magno-cellular (M-pathway) for low-frequency structures.

The process within each PMGB begins with an Adaptive Filter Unit (AFU). Given an intermediate feature $F_{in} \in \mathbb{R}^{C \times H \times W}$, we generate a content-adaptive low-pass kernel $\mathcal{K}_{LP}$ to decouple the components:
\begin{equation}
  F_L = F_{in} \circledast \mathcal{K}_{LP}, \quad F_H = F_{in} - F_L,
  \label{eq:freq_split}
\end{equation}
where $\circledast$ denotes the dynamic filtering operation. $F_L$ represents the Magno context, and $F_H$ isolates the Parvo residual high-frequency details. We then apply independent Self-Attention to both streams, yielding refined features $\hat{F}_L$ and $\hat{F}_H$.

In underexposed SDR inputs, the low-frequency background is often contaminated by quantization noise. Symmetric ``Coarse-to-Fine'' guidance may inadvertently propagate these artifacts. To prevent this, the PMGB enforces an Asymmetric Parvo-to-Magno Guidance ($H \rightarrow L$) flow. Once the high-frequency stream is processed and purified (via the SNN gating detailed in Sec.~\ref{subsec:snn}), $\hat{F}_H$ generates a channel-wise attention map to modulate $\hat{F}_L$:
\begin{equation}
  F_L^{*} = \hat{F}_L + \beta \cdot \big( \sigma(\mathcal{M}(\mathrm{AvgPool}(\hat{F}_H))) \odot \hat{F}_L \big).
  \label{eq:h2l_guidance}
\end{equation}
Here, $\odot$ denotes element-wise multiplication, $\mathcal{M}$ represents an MLP projection, and $\sigma$ is the Sigmoid activation. The learnable scaling factor $\beta$ ensures that high-frequency structural cues contribute to the low-frequency branch only when reliable edges are present.

To better model the receptive fields of simple cells in V1, $\hat{F}_H$ is further processed by an ON/OFF convolution parameterized as a Difference-of-Gaussians (DoG):
\begin{equation}
 \begin{aligned}
  W_{on}(x) &= A_c\,G(x;\sigma_c) - A_s\,G(x;\sigma_s),\\
  G(x;\sigma) &= \exp\!\Big(-\frac{\|x\|^2}{2\sigma^2}\Big),
  \label{eq:dog}
  \end{aligned}
\end{equation}
where $A_c,A_s>0$ and $\sigma_c < \sigma_s$. $W_{off}$ is set to $-W_{on}$ to maintain center-surround antagonism. This strict ON/OFF design encourages the Parvo branch to respond strongly to sharp discontinuities while suppressing broad low-frequency contamination.

\subsection{Event-Driven SNN Hard Gating} 
\label{subsec:snn}

A critical innovation within our PMGB is the purification of the high-frequency stream before it modulates the Magno pathway. Unlike conventional restoration frameworks that rely on explicit parameterized noise priors~\cite{chen2021hdrunet} or continuous ``soft gates'' such as Leaky ReLU, our proposed Event-Driven SNN Hard Gating module (Fig.~\ref{fig:snn_module}) acts as an implicit, non-parametric noise filter. As illustrated in Fig.~\ref{fig:snn_module}(a), traditional continuous activations lack a noise dead zone; they indiscriminately pass and entangle low-amplitude quantization noise, which is then severely amplified during HDR expansion. 

To address this, we leverage the temporal dynamics of spiking Integrate-and-Fire (IF) neurons (Fig.~\ref{fig:snn_module}(b)). Under IF dynamics, spatially-inconsistent quantization noise fails to drive sufficient membrane potential $V(t)$ accumulation to breach the learnable firing threshold $V_{th}$. Consequently, these artifacts are strictly trapped within the sub-threshold ``Noise Dead Zone'' and permanently blocked. The discrete-time IF dynamics are defined as:
\begin{equation}
  V(t+1) = \beta\,V(t) + F(t) - S(t)\cdot V_{th},
  \label{eq:if_dynamics_reduced}
\end{equation}
\begin{equation}
  S(t) = H\big(V(t) - V_{th}\big),
  \label{eq:spike_def_reduced}
\end{equation}
where \(F(t)\) is the input current, \(V(t)\) is the membrane potential, \(V_{th}>0\) is the threshold, \(\beta\in(0,1]\) is the leak, and \(H(\cdot)\) is the Heaviside step. We use a sigmoid-based surrogate for gradients during backpropagation:
\begin{equation}
  \tilde{S}(V) = \sigma\big(\alpha (V - V_{th})\big), \quad \frac{\partial S}{\partial V} \approx \sigma'\big(\alpha (V - V_{th})\big)\cdot \alpha.
  \label{eq:surrogate_grad_reduced}
\end{equation}
Spikes are averaged over \(T\) steps to form a normalized spike map \(S = \frac{1}{T}\sum_{t=1}^{T} S(t)\), which is then applied as a hard attention:
\begin{equation}
  F_{out} = F + \eta \cdot \mathrm{Conv}_{1\times1}(S).
  \label{eq:fout_reduced}
\end{equation}
To ensure the dead-zone functionality, we add a sparsity loss \(\mathcal{L}_{spar} = \| \bar{S} - r_{target} \|_1\) on the average firing rate. In single-image HDR reconstruction, weak noise struggles to accumulate stable potential over \(T\) steps, while continuous structural edges rapidly trigger spikes, elegantly separating details from artifacts.

\subsection{Objective Functions}
\label{subsec:loss}

To optimize Bio-SFT for both signal fidelity and perceptual quality, we employ a compound objective function. The total loss $\mathcal{L}_{total}$ is formulated as:
\begin{equation}
  \mathcal{L}_{total} = \mathcal{L}_{pix} + \lambda_{grad}\mathcal{L}_{grad} + \lambda_{spar}\mathcal{L}_{spar},
  \label{eq:loss_total}
\end{equation}
where $\mathcal{L}_{pix}$ denotes the standard pixel-wise reconstruction loss, and $\mathcal{L}_{grad}$ represents the gradient loss for high-frequency structural preservation. Crucially, $\mathcal{L}_{spar}$ is the sparsity loss applied to the Event-Driven SNN Hard Gating module, which explicitly penalizes the average firing rate to strictly maintain the sub-threshold noise dead zone. The hyperparameters $\lambda_{grad}=0.05$ and $\lambda_{spar}=0.01$ are empirically determined weights that balance the contributions of the gradient and sparsity penalties, respectively.

The pixel reconstruction loss minimizes the \(L_1\) distance between the reconstructed image \(I_{HDR}\) and the ground truth \(I_{GT}\):
\begin{equation}
  \mathcal{L}_{pix} = \| I_{HDR} - I_{GT} \|_1.
  \label{eq:loss_pix}
\end{equation}

The gradient fidelity loss penalizes mismatch in image gradients (implemented with Sobel kernels \(K_x,K_y\)):
\begin{equation}
  \mathcal{L}_{grad} = \sum_{\xi \in \{x, y\}} \| (I_{HDR} \ast K_{\xi}) - (I_{GT} \ast K_{\xi}) \|_1.
  \label{eq:loss_grad}
\end{equation}

\begin{table*}[t]
\centering
\caption{Quantitative comparison on the HDRTV1K dataset. The table highlights both the state-of-the-art performance of our standard architecture (\textbf{Bio-SFT}) and the extreme computational efficiency of our streamlined variant (\textbf{Bio-SFT-Lite}). Results with the best and second-best performance are highlighted in \textbf{bold} and \underline{underline}, respectively.}
\vspace{-2mm}
\label{tab:comparison}
\footnotesize
\resizebox{\textwidth}{!}{%
\begin{tabular}{lcccccccc}
\toprule
Method & Params (M) & PSNR $\uparrow$ & PU-PSNR $\uparrow$ & SSIM $\uparrow$ & PU-SSIM $\uparrow$ & SR-SIM $\uparrow$ & $\Delta E_{ITP} \downarrow$ & HDR-VDP-3 $\uparrow$ \\ \midrule
HuoPhyEO & - & 25.90 & 25.87 & 0.9296 & 0.9191 & 0.9981 & 38.06 & 7.893 \\
KovaleskiEO & - & 27.89 & 28.25 & 0.9273 & 0.9208 & 0.9809 & 28.00 & 7.431 \\
ResNet & 1.37 & 37.32 & 37.54 & 0.9720 & 0.9819 & 0.9950 & 9.02 & 8.391 \\
Pixel2Pixel & 11.38 & 25.80 & 26.16 & 0.8777 & 0.8901 & 0.9871 & 44.25 & 7.136 \\
CycleGAN & 11.38 & 21.33 & 21.54 & 0.8496 & 0.8319 & 0.9595 & 77.74 & 6.941 \\
HDRNet & 0.48 & 35.73 & 27.53 & 0.9664 & 0.9566 & 0.9957 & 11.52 & 8.462 \\
Ada-3DLUT & 0.59 & 36.22 & 36.17 & 0.9658 & 0.9781 & 0.9967 & 10.89 & 8.423 \\
Deep SR-ITM & 2.87 & 37.10 & 29.98 & 0.9686 & 0.9535 & 0.9950 & 9.24 & 8.233 \\
HDRTVNet & 1.41 & 37.61 & 37.74 & 0.9726 & 0.9820 & 0.9967 & 8.89 & 8.613 \\
DFT & \textbf{0.13} & 38.15 & 38.21 & 0.9737 & 0.9816 & 0.9967 & 8.25 & 8.586 \\
HDRTVNet++ & 0.59 & \underline{38.36} & \underline{38.39} & 0.9735 & 0.9820 & 0.9975 & \underline{8.28} & \underline{8.751} \\ 
\midrule
\textbf{Bio-SFT-Lite} & \underline{0.14} & 38.22 & 38.36 & \textbf{0.9743} & \textbf{0.9822} & \textbf{0.9976} & 8.25 & 8.745 \\
\textbf{Bio-SFT} & 0.48 & \textbf{38.53} & \textbf{38.43} & \underline{0.9740} & \underline{0.9820} & \textbf{0.9976} & \textbf{7.93} & \textbf{8.805} \\ \bottomrule
\end{tabular}%
}
\vspace{-4mm}
\end{table*}

\vspace{-1mm}
\section{Experiments}
\label{sec:experiment}

% Global settings to reduce space between floats and text
\setlength{\intextsep}{8pt plus 2pt minus 2pt}
\setlength{\textfloatsep}{8pt plus 2pt minus 2pt}

\subsection{Experimental Settings}
\label{subsec:settings}

\textbf{Datasets.} 
We conduct our experiments on two prominent benchmarks that represent different HDR generation paradigms: HDRTV1K~\cite{chen2021new} and SI-HDR~\cite{hanji2022sihdr}. 

\begin{figure}[b]
    \centering
    \includegraphics[width=\linewidth]{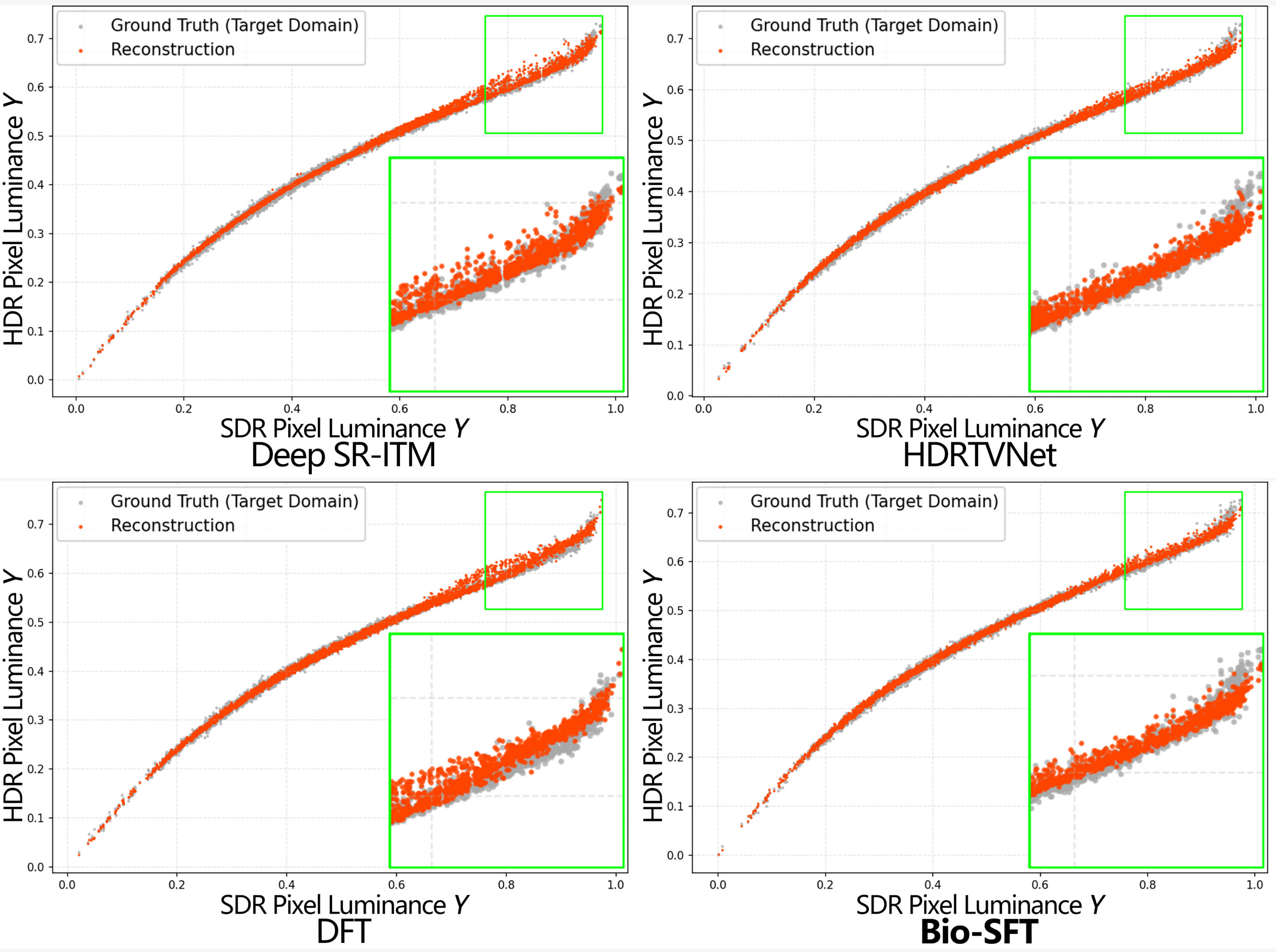}
    \vspace{-5mm}
    \caption{SDR-to-HDRTV pixel luminance mapping. The predicted curve follows the ground-truth trend under the EOTF-based evaluation scale.}
    \label{fig:luminance_mapping}
    \vspace{-3mm}
\end{figure}

Our primary training and evaluation focus on the HDRTV1K dataset, which aligns with the modern broadcast standards discussed previously. It contains 1,235 training pairs and 117 testing pairs of 4K resolution ($3840 \times 2160$) SDR-HDR images. Following prevalent broadcast-targeted benchmarks, we treat these high-resolution pairs as statistically independent frames during evaluation. During training, we randomly crop paired patches of size $96\times96$ without additional data augmentation. Because HDRTV1K complies with the ITU-R BT.2100 standard, we bypass explicit Camera Response Function (CRF) correction. Instead, we utilize the SMPTE ST 2084 Electro-Optical Transfer Function (EOTF) to convert predictions into absolute linear luminance ($cd/m^2$). As plotted in Fig.~\ref{fig:luminance_mapping}, this rigorous photometric alignment ensures our reconstructed trajectory tightly encloses the authentic ground-truth luminance curve.

Furthermore, to demonstrate that our proposed Bio-SFT is not strictly bound to PQ-encoded data but also generalizes robustly to traditional Inverse Tone Mapping (ITM) scenarios, we employ the SI-HDR dataset as a supplementary cross-dataset evaluation benchmark.

\textbf{Compared Baselines.} 
We benchmark our method against a comprehensive suite of state-of-the-art algorithms tailored to each dataset. For the primary HDRTV1K evaluation, we compare Bio-SFT with ten restoration pipelines under a standardized protocol: HuoPhyEO~\cite{huo2014physiological}, KovaleskiEO~\cite{kovaleski2014high}, ResNet~\cite{he2016deep}, pix2pix~\cite{isola2017image}, CycleGAN~\cite{zhu2017unpaired}, HDRNet~\cite{gharbi2017deep}, Ada-3DLUT~\cite{zeng2020learning}, Deep SR-ITM~\cite{kim2019deep}, DFT~\cite{xu2024dual}, and HDRTVNet~\cite{chen2021new}. For the cross-dataset evaluation on SI-HDR, we compare our framework against representative baselines, including classic CNN-based models (HDR-CNN~\cite{eilertsen2017hdr}, MaskHDR), pipeline-based methods (SingleHDR~\cite{liu2020single}), and recent deep architectures (Deep SR-ITM~\cite{kim2019deep}, DFT~\cite{xu2024dual}).

\textbf{Evaluation metrics.} We evaluate SDR-to-HDR reconstruction using seven full-reference metrics: PSNR, SSIM~\cite{wang2004image}, PU-PSNR and PU-SSIM (via PU21 encoding)~\cite{azimi2021pu21}, SR-SIM~\cite{zhang2012sr}, $\Delta E_{ITP}$ (ITU-R BT.2124), and HDR-VDP-3~\cite{mantiuk2023hdr}. PU-based metrics convert linear HDR samples into a perceptually uniform domain before applying conventional quality measures. This improves correlation with subjective HDR assessment. $\Delta E_{ITP}$ measures perceptual color difference in TV workflows. HDR-VDP-3 models the visibility of differences in HDR content and is computed with the same setup as in prior SDRTV-to-HDR work. Together, these metrics provide complementary views of reconstruction fidelity, perceptual consistency, and error visibility.

\begin{figure*}[b]
  \vspace{-3mm}
  \centering
  \includegraphics[width=\textwidth]{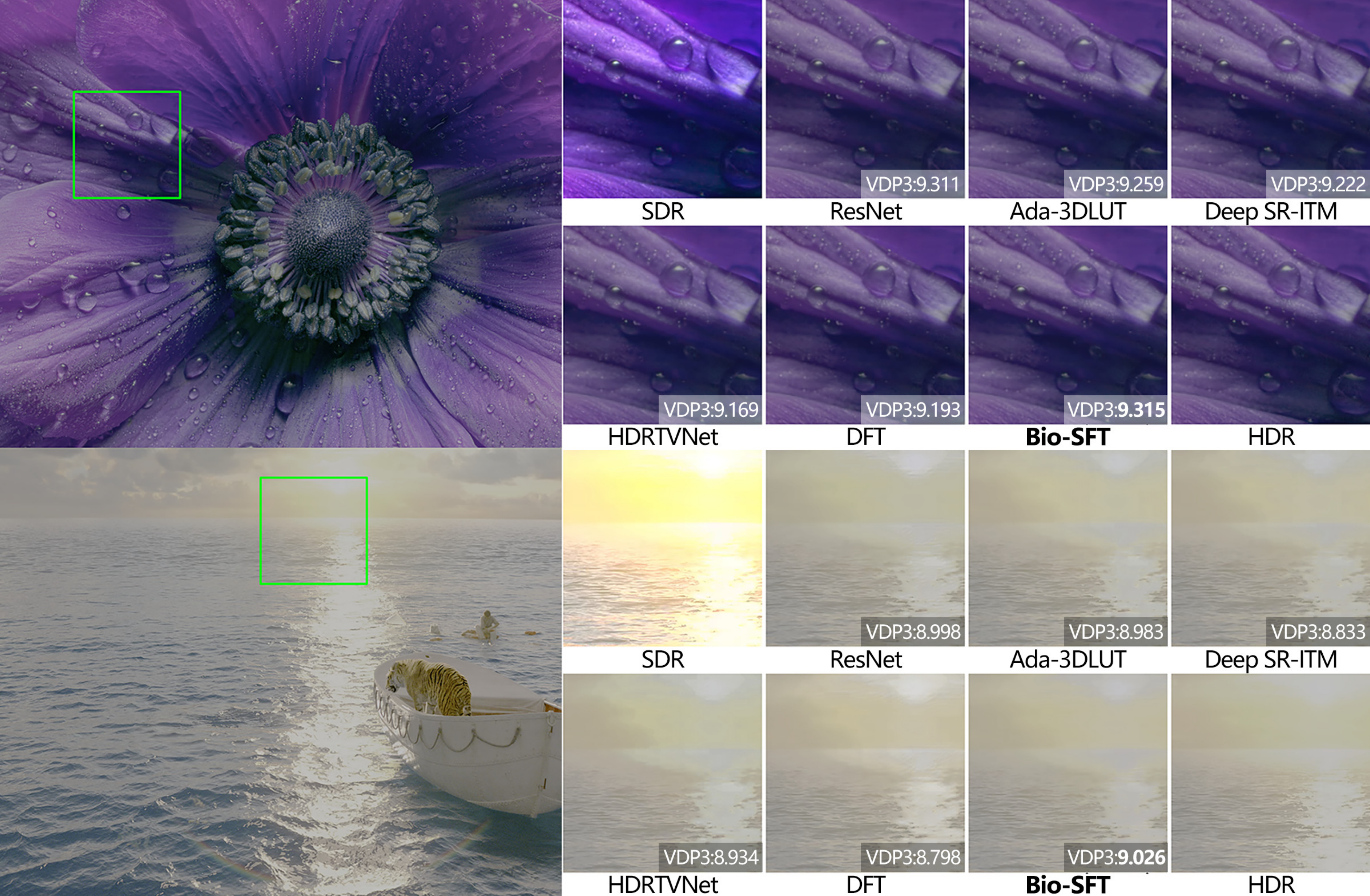}
  \vspace{-6mm}
  \caption{Qualitative comparison against state-of-the-art methods. Bio-SFT produces visually consistent colors and preserves fine structural details under challenging exposure transitions.}
  \label{fig:visual_results}
  \vspace{-3mm}
\end{figure*}

\textbf{Implementation details.} Bio-SFT is implemented in PyTorch and trained on NVIDIA RTX 3090 GPUs. We use Adam ($\beta_1=0.9, \beta_2=0.99$) with an initial learning rate of $7\times10^{-4}$ and a MultiStepLR scheduler. Training follows the progressive strategy described in Sec.~\ref{sec:method}. The same training protocol is used for all compared models whenever possible, so that performance differences are attributable to the model design rather than training settings.

\vspace{-2mm}
\subsection{Evaluation on HDRTV1K}
\label{subsec:hdrtv1k_eval}

\textbf{Quantitative Evaluation.} 
As outlined in our experimental settings, we benchmark our framework against state-of-the-art restoration pipelines. To demonstrate capacity ceiling and edge-device adaptability, we present two configurations: our standard architecture, \textbf{Bio-SFT}, and a streamlined variant, \textbf{Bio-SFT-Lite}. While Bio-SFT matches the parameter budget of modern architectures to maximize radiance accuracy, Bio-SFT-Lite is specifically engineered for resource-constrained deployment, drastically pruning redundancy while preserving the core physiological gating and asymmetric guidance mechanisms. 

As reported in Table~\ref{tab:comparison}, the standard Bio-SFT establishes a new state-of-the-art across the HDRTV1K benchmark. Compared to the highly competitive HDRTVNet++~\cite{chen2025towards}, Bio-SFT reduces the parameter burden by nearly 19\% (0.48M vs. 0.59M) while simultaneously yielding a PSNR gain of 0.17 dB and reducing the $\Delta E_{ITP}$ color error by over 4\%. This indicates a highly efficient utilization of network capacity for photometric accuracy. Furthermore, it achieves the highest HDR-VDP-3 score of 8.805, demonstrating superior perceptual alignment with the human visual system.

Remarkably, the streamlined Bio-SFT-Lite demonstrates extreme computational efficiency without sacrificing fidelity. Operating with a mere 0.14M parameters, it requires exactly 10\% of the parameter budget of conventional heavyweights like HDRTVNet (1.41M) and ResNet (1.37M), yet decisively outperforms them across all dynamic-range metrics. When evaluated against the similarly lightweight DFT (0.13M), Bio-SFT-Lite secures the top-tier structural similarity scores (an SSIM of 0.9743 and an SR-SIM of 0.9976) and a substantially higher HDR-VDP-3 response. The remarkably narrow performance gap between our two variants confirms that our bio-inspired asymmetric guidance and SNN gating---rather than brute-force parameter scaling---are the fundamental drivers of this optimal trade-off between computational efficiency and perceptual quality.

\textbf{Qualitative Evaluation.} 
To assess perceptual quality, we compare Bio-SFT with ResNet, Ada-3DLUT, Deep SR-ITM, HDRTVNet, and DFT. As shown in Fig.~\ref{fig:visual_results}, we select two representative scenes presenting distinct challenges: a macro shot of a purple flower (emphasizing texture and specular reflections) and a seascape sunset (emphasizing dynamic range and color gradients). These scenes effectively stress both local detail recovery and global tone consistency.

For the \textbf{purple flower scene} featuring water droplets on the petals, baseline methods exhibit various degradations. ResNet and HDRTVNet recover limited dynamic range, rendering the droplets unnaturally dim. Ada-3DLUT loses crucial details in shadow regions, obscuring the petal texture. Deep SR-ITM tends to produce overly saturated colors and harsh contrast, while DFT introduces a noticeable blue color cast in local regions. In contrast, Bio-SFT preserves realistic petal colors and meticulously retains structural details in both highlights and shadows, achieving a reconstruction closely aligned with the ground truth.

For the \textbf{seascape sunset scene}, where color fidelity and gradient smoothness are paramount, ResNet and HDRTVNet produce cooler and flatter tones. Deep SR-ITM artificially shifts the scene toward an orange tint, while Ada-3DLUT and DFT suffer from visible banding artifacts near the sun halo. Bio-SFT, governed by its asymmetric cortical guidance, maintains a highly consistent global tone. It preserves significantly smoother sky gradients and clearer transitions between bright specular reflections and surrounding shadows, directly reflecting our design goal of suppressing artifact propagation across large exposure variations.

\vspace{-1mm}
\begin{figure}[h]
  \centering
  \includegraphics[width=\linewidth]{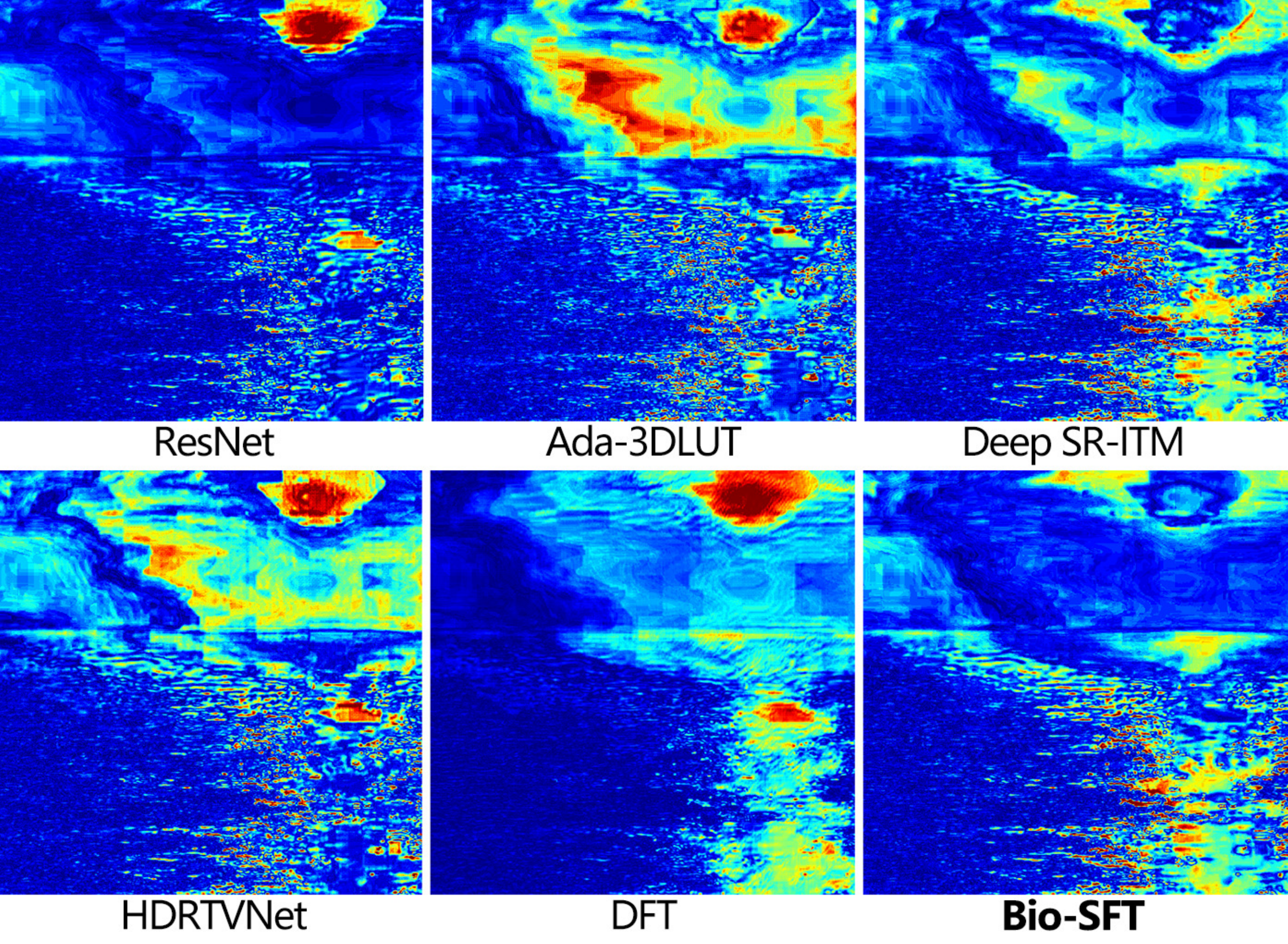}
  \vspace{-5mm}
  \caption{Pixel-wise absolute error maps visualized with a Jet colormap. Blue indicates low error and red indicates high error. Bio-SFT yields the lowest residuals in both textured regions and smooth gradients.}
  \label{fig:error_map}
  \vspace{-2mm}
\end{figure}

\textbf{Spatial Error Distribution.}
Fig.~\ref{fig:error_map} shows the pixel-wise absolute error maps. Bio-SFT produces the lowest reconstruction error, especially in high-frequency edge regions. This pattern is consistent with the bio-inspired design of the method, which explicitly filters sub-threshold artifacts. Early CNN-based architectures such as HDR-CNN~\cite{eilertsen2017hdr} and MaskHDR do not impose strong bio-physical constraints, so they may over-amplify brightness in saturated regions. For pipeline-based or parametric methods such as SingleHDR~\cite{liu2020single} and Ada-3DLUT~\cite{zeng2020learning}, the use of global response curves can lead to visible color shifts and banding in smooth regions. Symmetric frequency-fusion methods such as DFT~\cite{xu2024dual} may also allow noise from underexposed low-frequency backgrounds to affect high-frequency details, resulting in noticeable structural ghosting. By contrast, the proposed unidirectional Parvo-to-Magno guidance helps reduce such interference and produces more stable reconstructions. This does not imply that the alternative methods are fundamentally unsuitable; rather, it highlights the specific advantage of asymmetric guidance in the HDR setting.

\vspace{-2mm}
\begin{figure}[h]
    \centering
    \includegraphics[width=\linewidth]{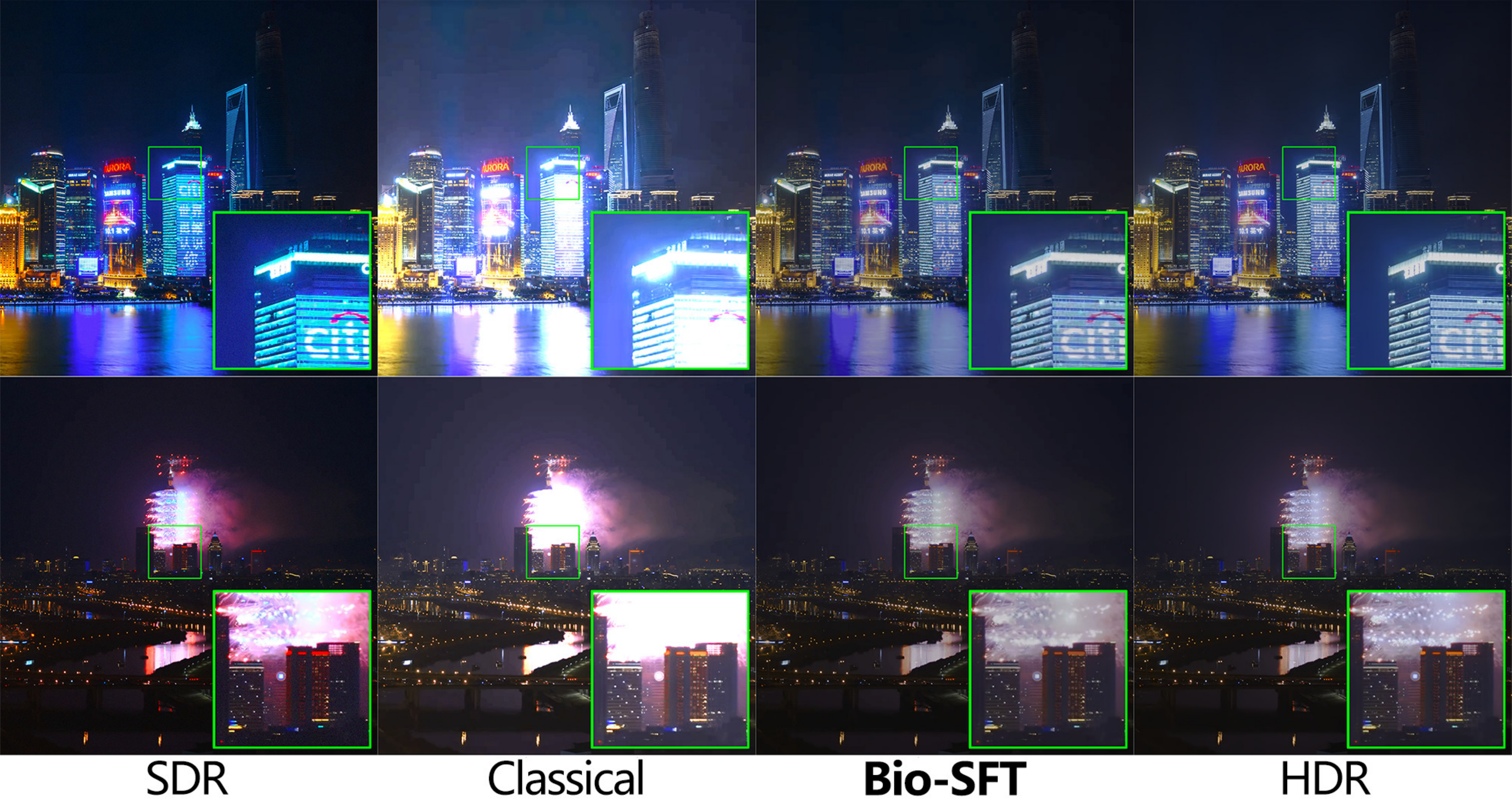}
    \vspace{-5mm}
    \caption{Visual comparison on an extremely illuminated scene. Bio-SFT reduces highlight overexposure while preserving structural details.}
    \label{fig:challenging_scene}
\end{figure}

\textbf{Challenging Scene Evaluation.}
Fig.~\ref{fig:challenging_scene} shows a highly illuminated scene containing strong specular reflections. Several baseline methods produce overexposed highlight regions because they amplify global brightness too aggressively without imposing local radiometric constraints. Bio-SFT successfully suppresses these clipping artifacts and recovers a physically plausible radiance distribution while effectively retaining fine texture. In this setting, the main advantage of the method is not only the reduction of clipping artifacts, but also the reliable preservation of local structural integrity around bright regions. This is important because the overall perceptual quality of HDR reconstruction often depends on how well the method handles such difficult, high-gradient luminance transitions without destroying adjacent chromatic information or introducing halo effects.

\begin{figure*}[t]
    \centering
    \includegraphics[width=\textwidth]{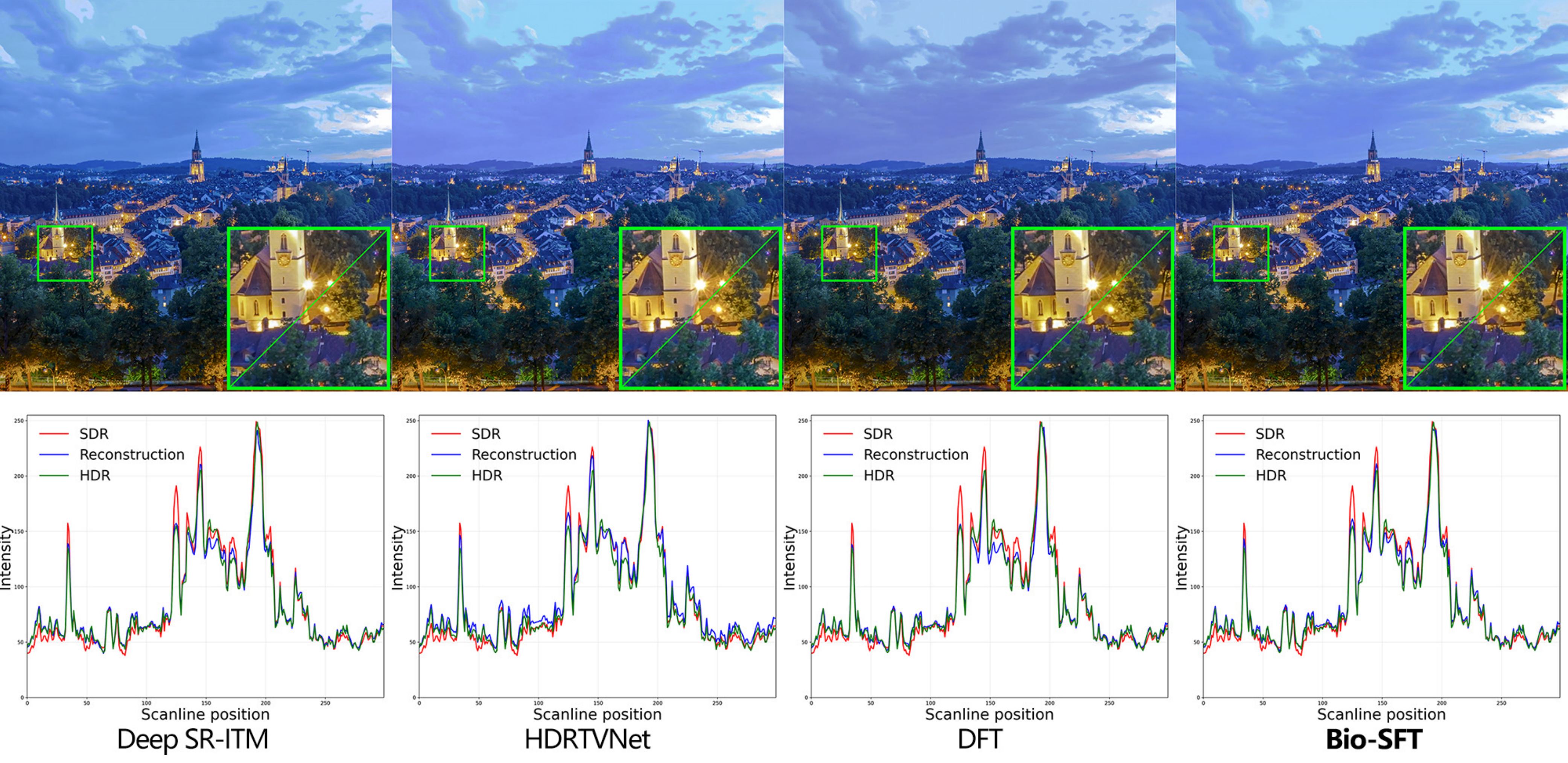}
    \vspace{-5mm}
    \caption{Scanline intensity profiles over saturated and dark regions. Bio-SFT follows the ground-truth intensity trend more closely.}
    \label{fig:scanlines}
    \vspace{-4mm}
\end{figure*}

\begin{figure*}[b]
    \vspace{-2mm}
    \centering
    \includegraphics[width=\textwidth]{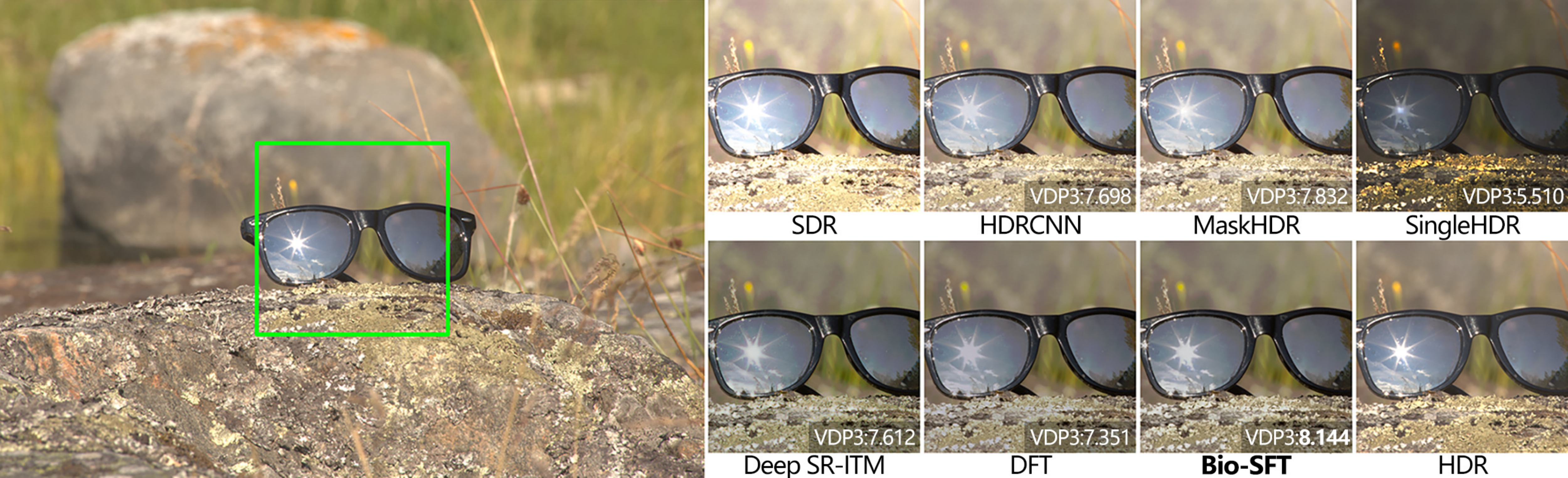}
    \vspace{-5mm}
    \caption{Visual comparison on the SI-HDR dataset (glasses scene). Bio-SFT uniquely avoids overexposure and color shifts, maximally preserving detailed lens structures and balanced contrast.}
    \label{fig:sihdr_glasses}
    \vspace{-4mm}
\end{figure*}

\textbf{Scanline Intensity Profiles.}
Fig.~\ref{fig:scanlines} presents scanline intensity profiles sampled across complex saturated and dark regions. Bio-SFT strictly follows the authentic ground-truth profile more closely than the comparison methods across the entire dynamic range. This suggests better radiance consistency across severe exposure transitions. The improvement is especially visible near steep luminance changes, where many baselines either overshoot, introducing severe ringing artifacts, or under-recover the true target intensity due to limited feature representation. The scanline plot therefore complements the visual comparison and error maps by empirically showing that the method preserves not only local high-frequency details but also the global linear luminance trend with exceptional photometric stability.

\subsection{Evaluation on SI-HDR}
\label{subsec:sihdr_evaluation}

\begin{table}[b]
    \vspace{-5mm}
    \centering
    \footnotesize 
    \setlength{\tabcolsep}{4pt} 
    \caption{Quantitative comparison on the SI-HDR dataset.}
    \vspace{-2mm}
    \label{tab:sihdr_results}
    \begin{tabular}{lcccc}
        \toprule
        Method & Params (M) & PU-PSNR$\uparrow$ & PU-SSIM$\uparrow$ & HDR-VDP-3$\uparrow$ \\
        \midrule
        HDRCNN & 29.44 & 25.71 & 0.9342 & 7.491 \\
        MaskHDR & 51.54 & 26.22 & 0.9390 & \underline{7.755} \\
        SingleHDR & 29.04 & \textbf{26.99} & \underline{0.9461} & 7.349 \\
        Deep SR-ITM & 2.87 & 25.93 & 0.9347 & 7.273 \\
        DFT & \textbf{0.13} & 25.82 & 0.9376 & 7.165 \\
        \midrule
        \textbf{Bio-SFT (Ours)} & \underline{0.14} & \underline{26.91} & \textbf{0.9482} & \textbf{7.893} \\
        \bottomrule
    \end{tabular}
\end{table}

\textbf{Quantitative Evaluation.} 
Table~\ref{tab:sihdr_results} summarizes the quantitative results on the SI-HDR benchmark. As shown in the table, Bio-SFT demonstrates remarkable cross-dataset generalization, achieving the highest structural similarity ($\text{PU-SSIM} = 0.9482$) and human perceptual fidelity ($\text{HDR-VDP-3} = 7.893$). While the pipeline-based SingleHDR attains a marginally higher PU-PSNR ($26.99\text{dB}$ vs. $26.91\text{dB}$), it requires over 200 times more parameters ($29.04\text{M}$) and suffers a severe drop in the HDR-VDP-3 perceptual metric. Furthermore, although recent architectures like Deep SR-ITM and DFT are relatively lightweight, they struggle to generalize on this novel data distribution, yielding significantly lower perceptual scores ($\text{HDR-VDP-3} < 7.3$). In contrast, our method maintains a superior balance, delivering state-of-the-art visual realism on an extremely lean parameter budget ($0.14\text{M}$).

\textbf{Qualitative Evaluation.} 
Fig.~\ref{fig:sihdr_glasses} presents a challenging scene featuring glasses to evaluate structural preservation and color rendering under complex illuminations. Early methods such as HDR-CNN and MaskHDR lack strict bio-physical bounds, causing them to blindly amplify global brightness and induce severe and unnatural overexposure in the background. SingleHDR struggles with accurate camera response reversal in this novel scene, resulting in a noticeable global color imbalance across the entire image. For modern deep architectures, Deep SR-ITM introduces harsh and unnatural color transitions around the lens area, while DFT suffers from a prominent yellowish color cast, likely due to low-frequency background noise leaking into the chromatic components. In contrast, Bio-SFT leverages the SNN hard gating and asymmetric cortical guidance to fundamentally block this frequency crosstalk. As a result, our method maximally preserves the fine structural details of the glasses frame and lenses while maintaining an authentic equilibrium of color and contrast, free from artificial color shifts or distracting saturation artifacts.

\subsection{Computational Efficiency at 4K Resolution.}
To assess deployment practicality, Table~\ref{tab:complexity} reports computational cost and inference time at 4K resolution ($3840\times2160$). Compared with HDRTVNet, Bio-SFT reduces MACs by $93.6\%$ and lowers runtime. It also uses fewer MACs than DFT while achieving higher reconstruction quality. These results indicate that Bio-SFT is suitable for hardware-constrained settings. At the same time, the runtime values should still be interpreted in the context of the specific hardware and implementation used in this study. In other words, the results support the efficiency of the design, but they do not by themselves guarantee the same runtime on all deployment platforms.

\begin{table}[h]
    \vspace{-2mm}
    \centering
    \footnotesize
    \caption{Complexity and efficiency comparison at 4K resolution.}
    \vspace{-1mm}
    \label{tab:complexity}
    \begin{tabular}{lcccc}
        \toprule
        Method & Params (M) & MACs (G) & Time (s) & PSNR (dB) \\
        \midrule
        HDRTVNet & 1.41 & 3112.99 & 3.09 & 37.61 \\
        DFT & \textbf{0.13} & 349.43 & 1.60 & \underline{38.15} \\
        \midrule
        Bio-SFT (Ours) & \underline{0.14} & \textbf{198.61} & \textbf{1.45} & \textbf{38.22} \\
        \bottomrule
    \end{tabular}
    \vspace{-3mm}
\end{table}

\subsection{Ablation Study}
\label{subsec:ablation}

We next analyze the contribution of each component. Tables~\ref{tab:ablation1}--\ref{tab:ablation3} summarize the main findings. The ablation results are useful because they separate the effect of each design choice from the full model behavior.

\textbf{Impact of Retinal Preprocessing.}
Table~\ref{tab:ablation1} compares linear input, conventional gamma correction, and the proposed bio-mimetic retina. Gamma correction reduces color fidelity and slightly lowers perceptual quality. In contrast, the learnable retinal front-end improves both color accuracy and HDR-VDP-3. This suggests that adaptive preprocessing is more suitable than fixed compression for HDR reconstruction. The improvement is not dramatic in every metric, but it is consistent across the most relevant perceptual measures.

\begin{table}[h!]
\vspace{-1mm}
\centering
\caption{Ablation 1:Input preprocessing.}
\label{tab:ablation1}
\vspace{-1mm}
\footnotesize
\setlength{\tabcolsep}{4pt}
\resizebox{\linewidth}{!}{%
\begin{tabular}{llccccc}
\toprule
ID & Preprocessing & PSNR $\uparrow$ & SSIM $\uparrow$ & SR-SIM $\uparrow$ & $\Delta E_{ITP} \downarrow$ & HDR-VDP-3 $\uparrow$ \\ \midrule
1-A & Linear / None & \underline{37.97} & \underline{0.9741} & \textbf{0.9976} & \underline{8.41} & \underline{8.525} \\
1-B & Gamma Correction & 37.75 & 0.9736 & 0.9972 & 8.68 & 8.410 \\
1-C & Bio-mimetic Retina (Ours) & \textbf{38.22} & \textbf{0.9743} & \textbf{0.9976} & \textbf{8.25} & \textbf{8.745} \\ \bottomrule
\end{tabular}%
}
\end{table}

\begin{table}[b]
\vspace{-1mm}
\centering
\caption{Ablation 2: High-frequency and SNN gating.}
\label{tab:ablation2}
\vspace{-1mm}
\footnotesize
\setlength{\tabcolsep}{4pt}
\resizebox{\linewidth}{!}{%
\begin{tabular}{lllccccc}
\toprule
ID & High-Freq Module & SNN & PSNR $\uparrow$ & SSIM $\uparrow$ & SR-SIM $\uparrow$ & $\Delta E_{ITP} \downarrow$ & HDR-VDP-3 $\uparrow$ \\ \midrule
2-A & Standard Conv & \XSolidBrush & 38.07 & 0.9739 & 0.9975 & 8.34 & 8.580 \\
2-B & ON/OFF Conv & \XSolidBrush & 37.95 & 0.9740 & 0.9975 & 8.42 & 8.560 \\
2-C & Standard Conv & \checkmark & \underline{38.15} & \underline{0.9742} & 0.9975 & \underline{8.30} & \underline{8.690} \\
2-D & ON/OFF Conv & \checkmark & \textbf{38.22} & \textbf{0.9743} & \textbf{0.9976} & \textbf{8.25} & \textbf{8.745} \\ \bottomrule
\end{tabular}%
}
\end{table}

\textbf{Synergy of High-frequency Modules and SNN Hard Gating.}
Table~\ref{tab:ablation2} studies the Parvo-pathway variants with and without SNN gating. The ON/OFF convolution alone slightly worsens color accuracy, which suggests that the filter is sensitive to high-frequency sensor noise in dark regions. When SNN gating is added, the low-amplitude noise is suppressed. The combination of ON/OFF convolution and SNN gating achieves the best overall results. This indicates that threshold-based sparsification complements edge-sensitive filtering. It also suggests that the high-frequency branch benefits from a more selective gating mechanism when the input contains strong quantization noise.

\begin{figure}[h]
    \centering
    \includegraphics[width=\linewidth]{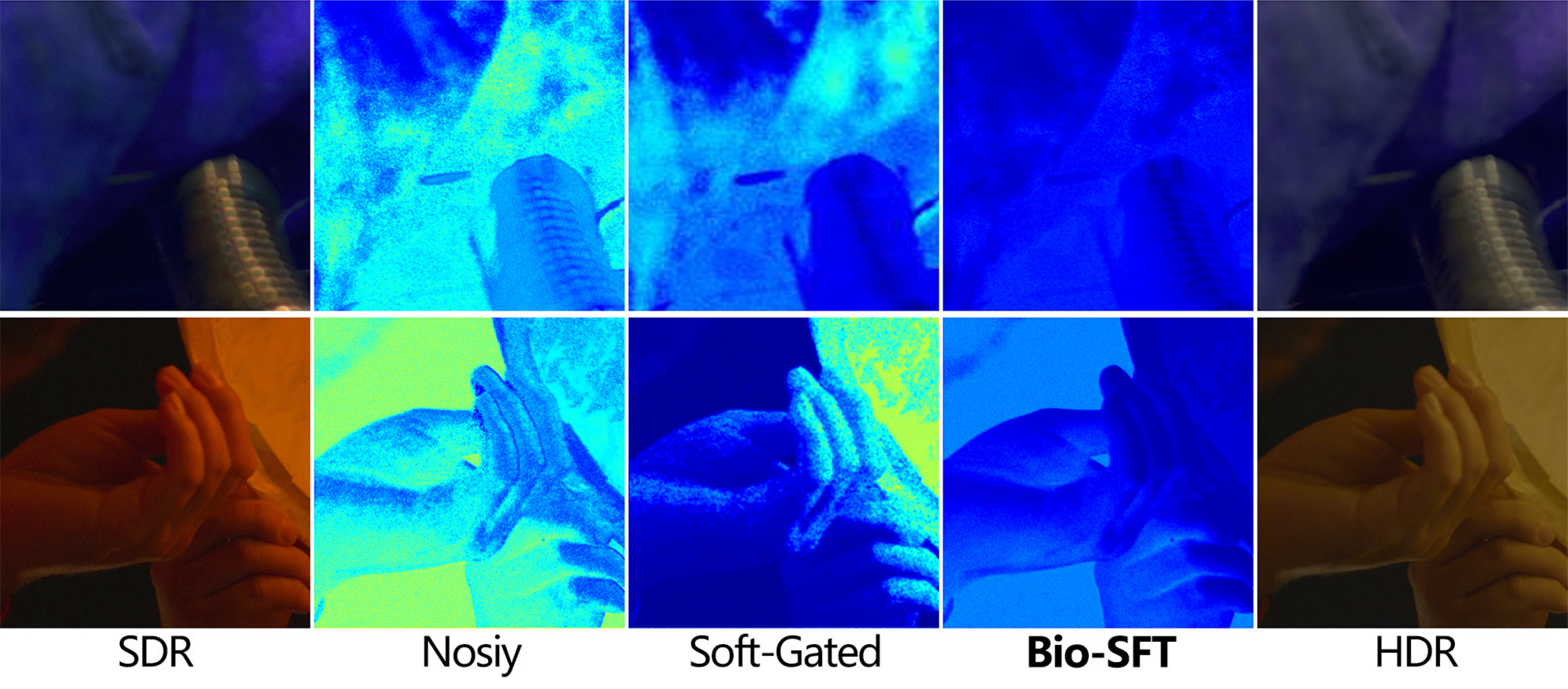}
    \vspace{-5mm}
    \caption{Channel-wise spatial activation maps in challenging dark scenes. Warm colors indicate high activation. The proposed SNN gating keeps background responses low while preserving sharp reflections.}
    \label{fig:activation_maps}
    \vspace{-2mm}
\end{figure}

Fig.~\ref{fig:activation_maps} shows the channel-wise spatial activation maps. Conventional continuous gating produces widespread activation in low-light backgrounds, which suggests that noise is not sufficiently suppressed. In contrast, the proposed SNN gating keeps the dark region more stable while preserving sharp reflections. This supports the role of hard gating in reducing weak fluctuations. The visualization does not prove strict denoising behavior in a formal sense, but it does provide qualitative evidence that the gating mechanism changes the activation pattern in a favorable way.

\textbf{Validation of Asymmetric Frequency Interaction.}
Table~\ref{tab:ablation3} compares different interaction strategies between low- and high-frequency streams. Bi-directional flow slightly increases color error and reduces PSNR, which suggests that low-frequency background signals may interfere with fine details. The proposed one-way Parvo$\rightarrow$Magno guidance yields the best trade-off between fidelity and compactness. This result is consistent with the central idea of the method: the more reliable high-frequency branch should guide the ambiguous low-frequency branch, rather than the reverse.

\begin{table}[h!]
\vspace{-1mm}
\centering
\caption{Ablation 3: Frequency interaction.}
\label{tab:ablation3}
\vspace{-1mm}
\footnotesize
\setlength{\tabcolsep}{4pt}
\resizebox{\linewidth}{!}{%
\begin{tabular}{llccccc}
\toprule
ID & Interaction Strategy & Params (M) & PSNR $\uparrow$ & SSIM $\uparrow$ & $\Delta E_{ITP} \downarrow$ & HDR-VDP-3 $\uparrow$ \\ \midrule
3-A & Summation / Concat & \underline{0.15} & \underline{38.01} & 0.9737 & \underline{8.38} & 8.582 \\
3-B & Bi-directional & 0.17 & 37.93 & \underline{0.9742} & 8.54 & \underline{8.653} \\
3-C & One-way (Ours) & \textbf{0.14} & \textbf{38.22} & \textbf{0.9743} & \textbf{8.25} & \textbf{8.745} \\ \bottomrule
\end{tabular}%
}
\end{table}

\section{Discussion}
\label{sec:discussion}

\textbf{Bridging Biological Plausibility and Artificial Efficiency.}
While traditional neuromorphic engineering relies heavily on specialized asynchronous hardware, Bio-SFT demonstrates that biological principles can be highly effective within standard synchronous deep learning frameworks. By abstracting the retinal adaptation and cortical gating into differentiable modules, we avoid the training instability typically associated with discrete spiking networks. This hybrid approach suggests a promising paradigm where biologically plausible mechanisms enhance, rather than replace, established deep learning operations.

\textbf{Mechanistic Synergy of the Visual Pathway.}
Beyond individual effectiveness, our architecture operates as a strict causal sequence. The retinal frontend must first stabilize the extreme dynamic range of the input; otherwise, the subsequent SNN thresholding would be overwhelmed by arbitrary illumination fluctuations. Furthermore, the SNN gating serves as a strict prerequisite for the asymmetric guidance: by explicitly trapping noise in the dead zone, it ensures that only pristine high-frequency structural cues dictate the modulation of the Magno stream. Thus, the performance gains stem from this tightly coupled physiological pipeline rather than isolated components.

\textbf{Algorithmic Limitations and Broader Impacts.}
Despite these advantages, the explicit thresholding in our SNN gating introduces hyperparameter sensitivity. The optimal dead-zone margin may vary depending on the inherent noise characteristics of different camera sensors, potentially requiring fine-tuning for completely unseen domains. Additionally, while the asymmetric guidance effectively suppresses spatial quantization artifacts in single images, handling complex spatio-temporal ghosting in multi-exposure HDR fusion remains an open challenge. Overcoming these hardware-specific and temporal challenges will be a valuable direction for future research.

\section{Conclusion}
\label{sec:conclusion}

In this paper, we presented \textbf{Bio-SFT}, a lightweight bio-inspired framework for single-image HDR reconstruction that combines a learnable retinal adaptation frontend, asymmetric Parvo-to-Magno guidance, and event-driven SNN hard gating. Unlike conventional symmetric coarse-to-fine designs, this processing flow effectively reduces noise propagation while preserving useful structural cues in signal-starved regions. Evaluations on both HDRTV1K and SI-HDR benchmarks confirm its effectiveness for perceptual quality and model efficiency. Specifically, the retinal frontend improves dynamic-range adaptation, the cortical guidance organizes frequency information flow, and the SNN gating provides sparse thresholding for weak responses, all of which are trained end-to-end within a hierarchical backbone.

While effective, the current framework should be viewed as a practical reconstruction model rather than a complete biological simulation. For instance, the present model focuses on single-image reconstruction, whereas video HDR requires additional temporal modeling; moreover, the spiking module functions primarily as a gating mechanism rather than a full neuromorphic inference system. Future work will explore temporal HDR video reconstruction and more explicit neuromorphic implementations. Ultimately, we expect that our overall design philosophy---combining adaptive preprocessing, frequency-aware guidance, and sparse gating---will benefit other efficient and robust low-level vision tasks.

\end{document}